\documentclass{article}

\usepackage{times}
\usepackage{soul}
\usepackage{microtype}
\usepackage{graphicx}
\usepackage{subfigure}
\usepackage{booktabs} 
\usepackage{color}
\usepackage{multirow}
\usepackage{hhline}
\usepackage{amsmath}
\usepackage{amssymb}

\usepackage{hyperref}

\newcommand{\iman}[1]{{\color{blue}Iman: #1}}

\newcommand{\todo}[1]{}

\usepackage{ijcai19}

\title{Multi-resolution Networks For Flexible Irregular Time Series Modeling (Multi-FIT)}

\author{
Bhanu Pratap Singh$^{1,}$\footnote{Equal contribution, randomly ordered.}
\and
Iman Deznabi$^{1,*}$
\and
Bharath Narasimhan$^1$
\and
Bryon Kucharski$^1$
\and\\
Rheeya Uppaal$^1$
\and
Akhila Josyula$^1$
\and
Madalina Fiterau$^1$
\\
\affiliations
$^1$College of Information and Computer Sciences\\
University of Massachusetts Amherst
\emails
\{brawat, iman, bnarasimhan, bkucharski, ruppaal, ajosyula, mfiterau\}@cs.umass.edu}

\begin{document}
\maketitle

\begin{abstract}

Missing values, irregularly collected samples, and multi-resolution signals commonly occur in multivariate time series data, making predictive tasks difficult. These challenges are especially prevalent in the healthcare domain, where patients’ vital signs and electronic records are collected at different frequencies and have occasionally missing information due to the imperfections in equipment or patient circumstances. Researchers have handled each of these issues differently, often handling missing data through mean value imputation and then using sequence models over the multivariate signals while ignoring the different resolution of signals. We propose a unified model named Multi-resolution Flexible Irregular Time series Network (Multi-FIT). The building block for Multi-FIT is the FIT network. The FIT network creates an informative dense representation at each time step using signal information such as last observed value, time difference since the last observed time stamp and overall mean for the signal. Vertical FIT (FIT-V) is a variant of FIT which also models the relationship between different temporal signals while creating the informative dense representations for the signal. The multi-FIT model uses multiple FIT networks for sets of signals with different resolutions, further facilitating the construction of flexible representations. Our model has three main contributions: a.) it does not impute values but rather creates informative representations to provide flexibility to the model for creating task-specific representations b.) it models the relationship between different signals in the form of support signals c.) it models different resolutions in parallel before merging them for the final prediction task. The FIT, FIT-V and Multi-FIT networks improve upon the state-of-the-art models for three predictive tasks, including the forecasting of patient survival.

\end{abstract}

\section{Introduction}
\label{sec:introduction}


Multivariate time series classification has significant applications, particularly in the medical, financial, automotive and networking domains where this type of data is abundant. In many cases however, despite the large volume of collected samples, multivariate time series present a unique set of challenges that make prediction difficult. PhysioNet \cite{Physionet} is one such dataset posing multiple difficulties. In the PhysioNet challenge, researchers try to predict the survival of the patients admitted to Intensive Care Units (ICUs) by modeling the vital signals collected over the span of their admission. Different vital signals such as heart rate and cholesterol levels are collected at different frequencies, posing the issue of \emph{multi-resolution} across vital signals (features). Vital signals such as heart rate are captured regularly but signals such as white blood cell count are captured \emph{irregularly}, depending on the patient's condition. 
The final issue faced by researchers is of \emph{missing data} which can occur because of device limitations. 

Significant effort has been expended in applying existing techniques and developing new ones to handle data with these characteristics, most significantly in the areas of Gaussian Processes~\cite{schulam2016disease}, functional clustering~\cite{yao2005functional,halilaj2018modeling} and deep learning~\cite{limdisease}. Nevertheless, these techniques are typically tailored to address one issue in particular, but fall short of providing a unified solution suitable for the classification of multi-resolution, irregularly sampled time series with missing data. As many datasets have all of these characteristics, the methods which are created to handle one of these problems would not be applicable without some form of data adaptation, incurring significant loss in performance in the process. 

Here we introduce the \emph{Flexible Irregular Time Series Network (FIT)}, which alleviates the shortcomings of prior work in classifying multivariate time series that present irregularities, signals of different frequencies, as well as missing values. Our method uses a \emph{memory cell} to compute informative representations for missing data rather than imputing a value, which are then passed on to an LSTM cell. The memory cell uses a fully connected neural network to better model trends in the temporal features, and is able to learn informative representations even with a limited number of observed values. This memory cell architecture also takes into account \emph{the time elapsed after the previous observation} when building a representation for a missing value. Further, we modified FIT to \emph{leverage the complex correlations between different features}, naming the resulting approach Vertical Flexible Irregular Time Series Network (FIT-V). FIT-V leverages the observations of other features along with past information for the feature requiring imputation. The main difference between the architecture of FIT model and previous approaches is that it doesn't perform value imputation for the signals at missing time steps but develops task-specific informative representations with the help of the global and local information of the signal. This provides FIT model a flexibility to retain task specific information rather than doing forced imputation which might add further noise to the sequence model.

The final model, called Multi-resolution Flexible Irregular Time Series Network (Multi-FIT), consists of two FIT components, one for slower signals and another for faster signals. Both the components operate parallely and are being fused prior to the prediction, thus jointly learning the final classification task. These components give model the power of handling data with different frequencies without needing to impute the slow signals at each time-step, which improves the predictive performance. We also enhanced the Multi-FIT model by using FIT-V as the base component to leverage the relation between different features, thus obtaining the Multi-FIT-V model.

Our approaches are specifically \emph{designed to be flexible} with respect to the input data, making them applicable to virtually any multivariate time series data, regardless of irregularity, frequency of collection or missing values. We demonstrate their superiority over the baseline model of mean-substitution combined with a Bidirectional LSTM with Attention, but also over the state-of-the-art models Temporal Belief Memory Network (TBM) \cite{kim2018temporal} and Gated Recurrent Unit with Decay (GRU-D) \cite{che2018recurrent} on real life datasets from the activity recognition and medical domains. Through experiments that remove some of the observations in a controlled manner, we studied the effects of missing data on the models, and illustrate that \emph{FIT is more robust than its contenders} in such scenarios. The experiments also indicate that FIT-V, and consequently Multi-FIT-V, further improves classification performance in the case when features are highly inter-correlated.

\section{Related Work}
\label{sec:related}

Previous work on multivariate time series classification functions under the assumption that the signals are aligned and collected at the same time steps, with missing values being imputed through simple methods such as mean-fitting and interpolation \cite{kreindler2016effects} or more complex procedures such as Expectation Maximization (EM) \cite{garcia2010pattern}, multiple imputation \cite{galimard2016multiple}, resampling \cite{cismondi2013missing} and kernel methods \cite{rehfeld2011comparison}. These methods are usually easy to implement and can yield good results when datasets have few missing values, but perform poorly when the dataset is too sparse \cite{che2018recurrent}.

Moreover, the missing values and their patterns can provide rich information regarding the dataset which cannot be captured by these methods \cite{little2014statistical}. In the case of an ICU patient, for instance, the frequency of clinical tests can also be indicative of the patient's current health status.
The features in multivariate time series datasets can have different complex distributions and simple imputation methods would not be able to model them accurately.
Although recent papers in this area offer solutions to datasets presenting these challenges, they usually focus on solving one of the aforementioned problems. For instance, \cite{che2018recurrent} try to leverage patterns of missing values in recurrent neural networks to improve the prediction performance, while \cite{li2016scalable,li2015classification} attempt to solve the problem of irregularity in uni-variate time series data. However, none of these methods focus on offering a flexible design to 
handle all of the outlined issues, ubiquitous in medical datasets.

More recent models such as Temporal Belief Memory network (TBM) \cite{kim2018temporal} offer solutions to solve these problems. It is a bio-inspired method which imputes data before feeding it into a sequence model. It uses a decaying formula for imputing data according to the last observed and average value of the signal. Thus, the method falls short in considering the relationship between the features and providing a flexible model for complicated patterns. Gated Recurrent Unit with decay (GRU-D) \cite{che2018recurrent} utilizes missing patterns (informative missingness) in medical data for effective imputation and for improving the predictive performance of the sequence models. Although GRU-D provides a flexible model to capture complex patterns, it does not use information from the neighboring features and fails to model multi-resolution properties which are often encountered in the medical domain. Finally, though all of these methods employ deep sequential models to exploit the temporal information of the data, none of them use attention \cite{luong2015effective}, which was introduced in sequence networks to enable \emph{focus} on different spans of time series data before making the final prediction.

\begin{figure*}[t]
\centering
  \includegraphics[width=17cm]{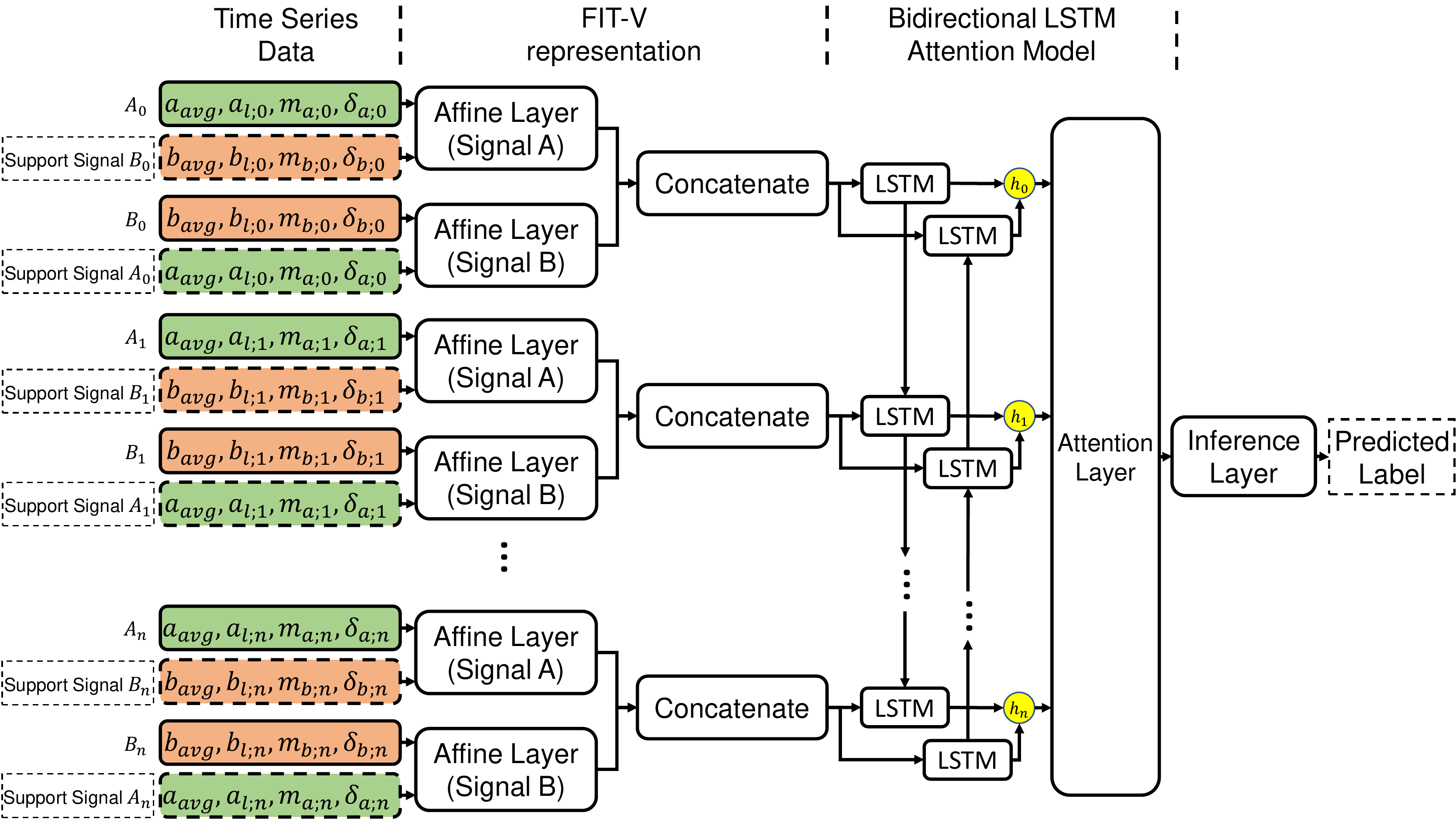}
  \caption{Vertical Flexible Irregular Time Series Network (FIT-V). An example of time series data with 2 signals $a$ and $b$ have been shown. Both signals act as support feature for each other. Their vectors ($[a_{avg}, a_{l;i}, m_{a;i}, \delta_{a;i}]$) pass through an affine layer to build a representation. The representation of the signal and it's support is passed through a sequence network with global attention before passing through inference layer for prediction.}
  \label{dbm_fig}
\end{figure*}

\section{Methods}
\label{sec:Methods}
\subsection{Problem Setup}
Multivariate time series data suffers mainly from three problems: \textit{irregularity}, \textit{missingness} and \textit{multi-resolution signals}. In the medical domain these problems are quite prevalent. Assume that for a patient, there are $M$ medical signals which could be extracted from their electronic health records. For signal $i$, where $i\in\{1...M\}$, we have the set of observations $x_{i;1}...x_{i,n_i}$, available at time stamps $t_{i;1}....t_{i,n_i}$, where $n_i$ is the number of observations for the signal $i$ and can differ widely over signals. Notably, values for any two signals can be collected independent of each other or could not be noted precisely in the clinical note. Similarly, at times there can be two vital signals which are collected for the patient: $x_{slow}$ and $x_{fast}$ where $x_{slow}$ is collected at lower frequency than $x_{fast}$, for each observation $x_{slow;i}$ there could be several observations $x_{fast;r^1_i}...x_{fast;r^2_i}$, where the indices $r^1_i$ and $r^2_i$ indicate the relevant part of the fast signal for the $i^{th}$ value in the slow signal. This characteristic makes it difficult to align the different signals to the same time steps, a prerequisite to applying a generic sequence network. In order to tackle these problems, we discuss our proposed techniques in this section.

Most of the existing techniques are limited in their capability to capture the complex patterns of the signals for imputing the time series data. They tend to use specific functions or rule-based methods \cite{donders2006gentle} to impute the values for different signals. Rule-based methods can be quite effective at times but require in-depth domain knowledge. Also, constructing rules for each of the signals in a dataset with a large feature set could prove to be tedious. These techniques also fail to leverage the information provided by the other features while imputing values for a signal. High signal correlation is another frequent characteristic of medical data, and can be used to improve imputation accuracy.

In order to compare our model against other methods, which perform imputation, we coupled those methods with a sequence model similar to our architecture.
For the sequence model we selected a Bidirectional Long Short-Term Memory network with global attention (BiLSTM Attn). BiLSTM Attn along with FIT, FIT-V and MultiFIT techniques that handle the problems of irregular, missing and multi-resolution data, are explained below. 

\subsection{Bidirectional LSTM with Attention (BiLSTM Attn)} 
\label{sec:bilstm_attn}
The Bidirectional Long Short-Term Memory (BiLSTM) network has two LSTM layers \cite{hochreiter1997long} where the first layer propagates in the forward direction and the second propagates backwards. The hidden states from both forward and backward LSTM are concatenated to form the final hidden state for the BiLSTM model.
\begin{equation}
\overrightarrow{LSTM}(x_t)+  \overleftarrow{LSTM}(x_t) = \overrightarrow{h_t} + \overleftarrow{h_t} = h_t
\end{equation}

Where $x_t$ is the input signal to the BiLSTM, this would be different for each variant of our model. The BiLSTM uses the signals at each time step to obtain the corresponding hidden representation.
\begin{equation}
    H = [h_{1}, h_{2} ....h_{t}... h_{n}] 
\end{equation}
The hidden representation for the whole time series ($H$) is then provided as input to an affine layer ($W_a$) and a softmax layer to get the location based global attention \cite{luong2015effective}. The concept of attention is common in natural language processing where it helps emphasize different tokens in a sequence of words. Attention helps the sequence model in focusing on important spans of the time series sequence.
\begin{equation}
\label{equ_at}
    a_t = softmax(W_ah_t)
\end{equation}
This attention is further multiplied with $h_t$ to get the final hidden representation ($h'_t$) for the time series
\begin{equation}
\label{equ_ht}
    h'_t = a_th_t
\end{equation}
This final hidden representation ($h'_t$) is used to predict the label for the time series. 

\subsection{Flexible Irregular Time Series Network (FIT)}         %
We propose Flexible Irregular Time Series (FIT) networks, which can encode the different complex patterns present in multivariate time series. FIT takes four values as input at each time step for each signal $x$: a missing flag ($m_{x;t}$), the average value of the signal ($x_{avg}$), the last observed value of the signal ($x_{l;t}$) and the time lapse from the last observed value ($\delta_{x;t}$).

FIT defines an affine layer for each signal of the time series data which takes a vector as an input containing the four values defined above: $[x_{avg}, x_{l;t}, m_{x;t}, \delta_{x;t}]$. A separate affine layer serves two purposes: \textit{first}, it gives the network the flexibility to model complex patterns for each signal separately and \textit{secondly}, it generates a representation for the signal at each time step rather than performing a direct value imputation which allows the model to have the flexibility of retaining only the task-specific information. These representations also enable model the \textit{flexibility} to learn different signal patterns before performing the final label prediction. These separate affine layers act as \emph{memory cells} for the features which help in getting a vector representation at each time step.

First, a vector is created for a signal $a$ at time $t$: 
\begin{equation}
    V_{a;t} = [a_{avg}, a_{l;t}, m_{a;t}, \delta_{a;t}]
\end{equation}
$V_{a;t}$ is then provided as input to an affine layer ($W_a$) to get a representation for this signal, as follows 
\begin{equation}
    R_{a;t} = W_aV_{a;t}
\end{equation}
Assuming we are operating on a multivariate time series with signals $a$, $b$ and $c$, the final representation is created by concatenating the signal representation.
\begin{equation}
    R_t = [R_{a;t}, R_{b;t}, R_{c;t}]
\end{equation}
This time series of $R=[R_1...R_n]$ is then provided to a BiLSTM Attn model, as explained in section~\ref{sec:bilstm_attn}.

At each time step, the signal vector is passed through its memory cell to get a representation. The representations for all signals are concatenated in the next layer and are provided to the cell of the sequence network to obtain the hidden representations ($h_t$). These hidden representations are then used by the attention layer to get the final sequence representation for the whole time series, which is used by the inference layer to get the label prediction. 
\subsection{Vertical Flexible Irregular Time Series Network (FIT-V)}
Next, we introduce the Vertical Flexible Irregular Time Series (FIT-V), a variant of the FIT network. While generating a representation for a signal at a particular time step, FIT-V also looks at information vertically, taking into account the values of other signals at the same time step, as signal correlation can provide important information regarding missing values. In FIT-V, we define \emph{support signals} for each signal that help construct its representation. In all our experiments, we provide \emph{support signals} for each signal according to the inter-correlation between the signals calculated over training data. The correlated support signals would help in building meaningful representation of the missing signal.

Suppose we have three signals: $a$, $b$ and $c$ and each has one support signal. The support signal for $a$ is $b$, for $b$ it is $c$ and for $c$ it is $a$. Then both vectors are concatenated and provided as input to the \emph{memory cell} (affine layer, $W_a$).
\begin{equation}
    R_{a;t} = W_a[V_{a;t}, V_{b;t}]
\end{equation}
Similarly, we can calculate $R_{b;t}$ and $R_{c;t}$ and eventually $R_t = [R_{a;t}, R_{b;t}, R_{c;t}]$. This time series of $R=[R_1...R_n]$ is then used by the BiLSTM model with global attention as explained in section~\ref{sec:bilstm_attn} for the final label prediction.

In medical datasets, vital signals such as heart rate and blood pressure are highly correlated. In such cases, FIT-V, shown in Figure~\ref{dbm_fig}, would be quite helpful as it leverages the information from correlated signals while generating the representation for a particular signal. According to domain knowledge, the user could define a set of additional signals to be used for each signal while generating its representation. 

\begin{figure}[!htbp]
\centering
  \includegraphics[width=8cm]{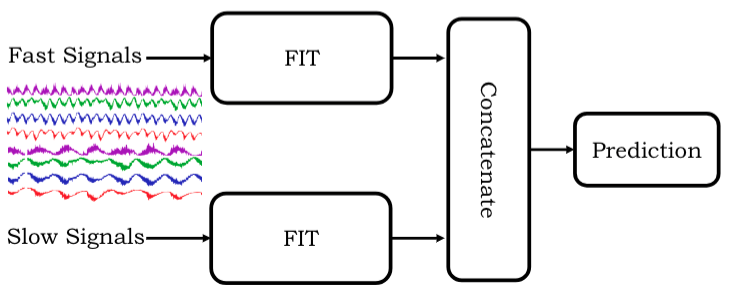}
  \caption{MultiFIT architecture.}
  \label{multiFIT}
\end{figure}

\subsection{Multi-Resolution FIT (multi-FIT) and Multi-Resolution FIT-V (multi-FIT-V)}
The Multi-Resolution FIT (Multi-FIT) and Multi-Resolution FIT-V (Multi-FIT-V) models were designed to better handle the case of varying resolutions across signals in multivariate time series. Multi-FIT splits the signals into dedicated branches each with its own FIT network, trained to handle data of a particular frequency range. The \emph{fast} (\emph{slow}) branch of each model is given only \emph{high} (\emph{low}) frequency data, thus allowing each branch to learn fine grained information from its data with more ease. Multi-FIT-V operates the same way, with the notable distinction that it uses FIT-V as opposed to FIT networks.


Each signal of every time series instance is classified as either a \emph{fast} signal or a \emph{slow} signal. 
To generate this classification, we calculated the frequency of each signal based on $\delta_t$ values which represent the time passed from the last observed value of the signal at each time step.
Signals with lower frequency are classified as slow, and those with higher frequencies are classified as fast. The higher the summation value of $\delta_t$ for the signal, the slower it would be\footnote{A clear distinction between the \textit{fast} and \textit{slow} signals can be observed with the help of Fig.4 presented in Appendix A.}.

Each branch of the Multi-FIT network consists of memory cells over all features, followed by a bidirectional LSTM, followed by an attention layer. These representations from the `fast' and `slow' branches are then concatenated and provided to a final set of affine layer which generate the class label. Similarly, the Multi-FIT-V branches consist of the memory cells over all features and their \textit{support features}, followed by the Bidirectional LSTM and attention mechanism. The model can be divided into multiple frequency branches, we chose two branches because our signals could be divided clearly into two groups as shown in Appendix A.

\section{Experiments and Results}
\label{sec:Experiments}

To evaluate our proposed approaches - FIT, FIT-V, multi-FIT, and multi-FIT-V, we conducted a series of experiments on three real datasets, namely, the Inertial Sensor Dataset by Osaka University (OU-ISIR), the Human Motion Primitives Detection (HMP) dataset and the PhysioNet dataset. These datasets are  popular benchmarks for medical time series modeling, as indicated by the related work. For each dataset, we compare our models against a baseline which first performs mean value imputation then provides the mean imputed time series data to a BiLSTM Attn model, we refer to this baseline model as BiLSTM model with Attention over mean imputed data (BA-Mean). Additionally, our models are evaluated against different state-of-the-art models such as TBM for OU-ISIR and GRU-D for PhysioNet which are commonly used for such irregular multivariate time series classification.

For the OU-ISIR and HMP datasets, we randomly removed a fixed amount of data for each signal to perform a controlled experiment of how well the models handle missing values. For these datasets, we also provide the results of BiLSTM Attn obtained prior to the removal of the values. We refer to this as BA-Oracle, and it represents an upper bound on the performance of methods we tested. When there is no missing data, TBM performs identically to BiLSTM Attn. Thus, on the full dataset, the performance of TBM will be the same as that of BA-Oracle. GRU-D is designed to leverage patterns of missing data, and thus we compare against it on the PhysioNet challenge, where such patterns exist.

Each dataset is divided into training, validation and test sets where the hyperparameters are selected over the validation set. The \textit{precision}, \textit{recall} and \textit{f-score} values achieved by each contender over the test set are provided in Table~\ref{table:prf_table}. The features and processing steps of each dataset along with the results achieved are explained in the following sub-sections. 


\subsection{OU-ISIR Gait Database, Inertial Sensor Dataset}
The OU-ISIR Gait Database, Inertial Sensor Dataset (OU-ISIR)  \cite{ngo2014largest} is a dataset on inertial sensor-based gait. During data collection, each subject wore 3 IMUZ sensors and a phone around their waist. The IMUZ sensors recorded 3D gyroscope and 3D accelerometer values, while the phone only recorded 3D accelerometer values. Altogether, these sensors recorded 21 features. The predictive task associated with this dataset is activity recognition, i.e., determining one of the four tasks of level walk for (1) entering and (2) exiting, (3) moving up-slope and (4) moving down-slope being performed, based on sensor readings.

The two level walks are very similar, as they both consist of walking on a flat surface. Therefore, for our experiments, the two walking labels were merged. To evaluate the imputing power of our models, 60\% of the observed values were randomly removed. We obtained what we expected to be an upper bound on performance by learning a BiLSTM-Attn network without any missing data, which is referred to as BA-Oracle in Table \ref{table:prf_table}. The BiLSTM Attn network where missing value is replaced by mean value of remaining signals acted as the baseline contender and lower bound on performance. This model is referred as BA-mean. The data was divided into training, validation and testing set, in a ratio of 64:16:20. 

The first row of Table~\ref{table:prf_table} summarizes these results. As expected, BA-Oracle achieves the highest performance, with an F-score of $0.963$ on the test set. The best model out of the evaluated models is FIT-V, which achieves an f-score of $0.924$. FIT's performance (f-score: $0.907$) is close to that of FIT-V. For FIT-V, each feature was provided with 5 \emph{support} features. These 5 features were selected as \emph{support signals} on the basis of their inter-correlation with the feature in the training set. The results are compared against Temporary Belief Memory (TBM) network as it is a state-of-the-art for irregularly missing time series data, it achieves an f-score of $0.873$ which is only slightly better than BA-mean. Thus, FIT-V performs significantly better than the baseline as well as the state-of-the-art model.

As FIT-V was designed for the case when there is significant correlation between features, we conducted an additional experiment to verify that it actually performs as expected in such a scenario. We created another dataset called OU-ISIR-Corr, by selecting the 5 most correlated features amid the 21 features available in OU-ISIR, as determined by computing pairwise Pearson correlation coefficients. Each feature was then provided with remaining 4 features as \emph{support signals} within FIT-V. For OU-ISIR-Corr, the BA-Oracle model achieves an f-score of $0.960$, which is quite close to BA-Oracle's performance on the original OU-ISIR dataset. The model which is closest in performance to BA-Oracle for OU-ISIR-Corr is FIT-V, with an f-score of $0.856$ which is higher than FIT's achieved f-score of $0.847$. These results shows two things: (1) FIT-V consistently performs better than its counterpart FIT when the features are inter-correlated and (2) when the number of features are reduced, the FIT models experience lesser drop in performance as compared to TBM and BA-mean.

\begin{figure}[t]
    \centering
    \includegraphics[width=0.48\textwidth]{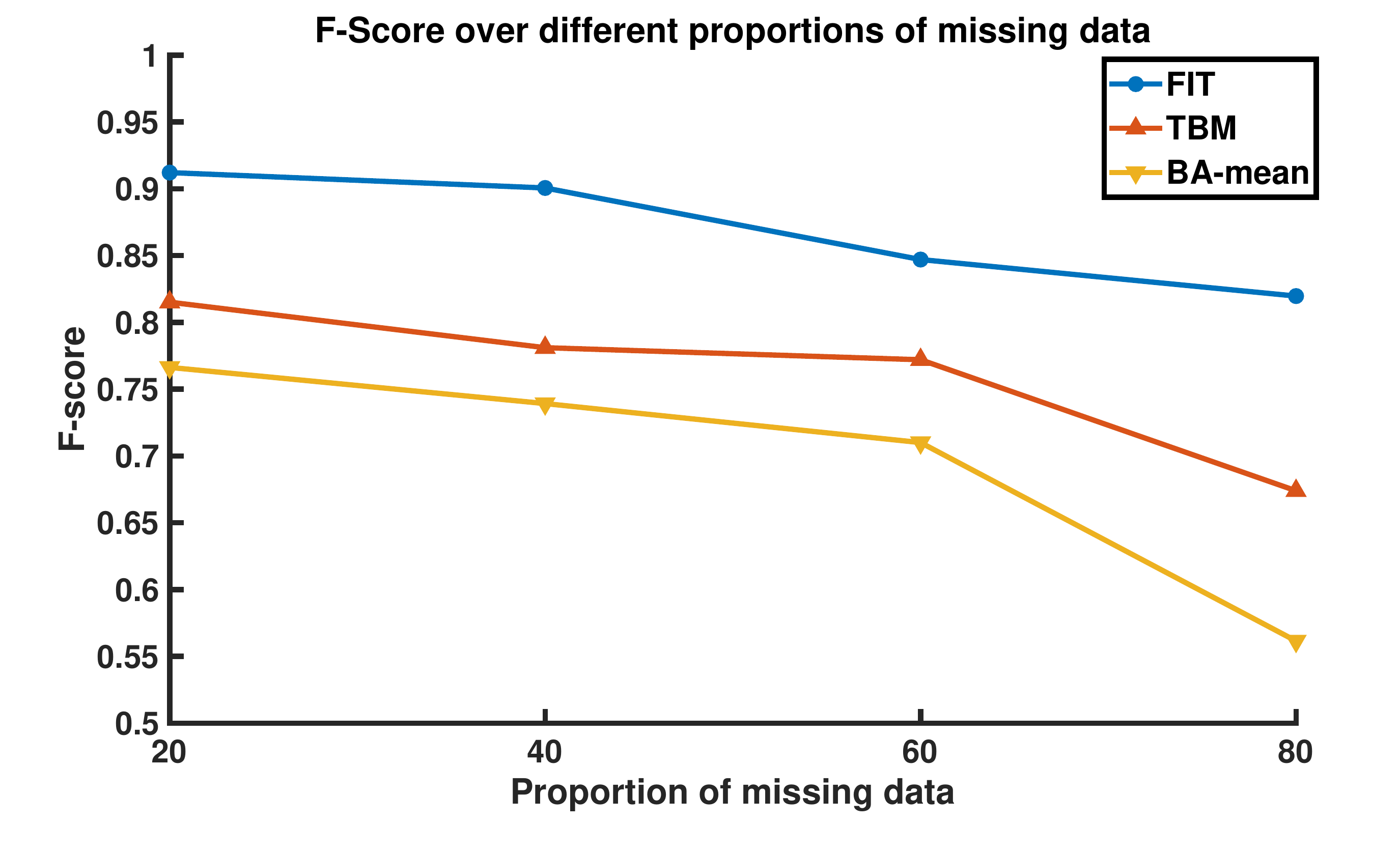}
    \caption{The performance of FIT, TBM and BA-mean on OU-ISIR-Corr dataset with different portion of missing values.}
    \label{fig:missingProp}
\end{figure}

To check the robustness of FIT models, we ran an experiment where we increased the amount of missing data and noted the change in performance. As shown in Figure~\ref{fig:missingProp}, FIT consistently outperforms its contenders, TBM and BA-mean, showing its robustness to large amounts of missing data.


\subsection{Dataset of Accelerometer Data for Human Motion Primitives Detection (HMP)}
The dataset of Accelerometer Data for Human Motion Primitives Detection (HMP) \cite{bruno2012human} consists of accelerometer data collected from 16 volunteers doing 14 activities of daily life via a single tri-axial accelerometer attached to the right wrist of the volunteer. The dataset consists of three values $(X, Y, Z)$ recorded over time while the volunteer does any of the 14 tasks. Here again, we randomly removed 60\% of the data to assess the effectiveness of our methods. 
The goal for this application is predicting the subject's activity from the time series data recorded via the sensor. 

This dataset was also divided into training, validation and test sets in a ratio of 64:16:20. The best and worst performing models were BA-Oracle and BA-mean, respectively, achieving F-scores of $0.831$ and $0.705$. As shown in the third line of Table~\ref{table:prf_table}, FIT-V achieved the closest performance to BA-Oracle with an F-score of $0.802$. Once again, FIT-V ($0.802$) and FIT ($0.785$) achieve better performance than TBM and BA-mean. In HMP dataset, the three features are quite correlated with each other which is why for each feature the other two features act as \emph{support signals} for FIT-V model.
%

\begin{table}[!h]
\renewcommand{\arraystretch}{1}
\centering
\small
\begin{tabular}{l|l|c|c|c}
\hline
\textbf{Datasets}                      & \textbf{Models}    &  \textbf{Precision} & \textbf{Recall} & \textbf{F-score} \\ \hline \hline
\multirow{5}{*}{\textbf{OU-ISIR}}      & \textbf{BA-Oracle} & 0.963              & 0.963           & 0.963            \\ 
\hhline{~----} 
                                       & \textbf{FIT-V}      & 0.938              & 0.910           & \textbf{0.924}            \\ 
                                       & \textbf{FIT}       & 0.910              & 0.907          & 0.907   \\ 
                                       & \textbf{TBM}       & 0.882              & 0.874           & 0.873            \\ 
                                       & \textbf{BA-mean}    & 0.770              & 0.948           & 0.850            \\ \hline
\multirow{5}{*}{\textbf{OU-ISIR-Corr}} & \textbf{BA-Oracle} & 0.960              & 0.960           & 0.960            \\ \hhline{~----} 
                                     & \textbf{FIT-V}        & 0.864     & 0.855  & \textbf{0.856}   \\ 
                                     & \textbf{FIT}         & 0.853              & 0.843           & 0.847            \\ 
                                     & \textbf{TBM}         & 0.772              & 0.772           & 0.772            \\ 
                                     & \textbf{BA-mean}      & 0.719              & 0.710           & 0.710            \\ \hline
\multirow{5}{*}{\textbf{HMP}}        & \textbf{BA-Oracle}   & 0.847              & 0.833           & 0.831            \\ \hhline{~----}
                                     & \textbf{FIT-V}        & 0.823     & 0.815  & \textbf{0.802}            \\ 
                                     & \textbf{FIT}         & 0.792              & 0.798           & 0.785            \\ 
                                     & \textbf{TBM}         & 0.807              & 0.780           & 0.772            \\ 
                                     & \textbf{BA-mean}      & 0.701              & 0.726           & 0.705            \\ \hline
\multirow{7}{*}{\textbf{PhysioNet}}  & \textbf{Multi-FIT-V} 
& 0.681  &	0.690 &	\textbf{0.681}
\\
                                    & \textbf{Multi-FIT} 
                                    & 0.668 &	0.676 &	0.663
                                    \\
                                    & \textbf{FIT-V}     
                                    
                                    & 0.684 &	0.685 &	0.676
                                    \\ 
                                     & \textbf{FIT}      
                                     & 0.666 &	0.676 &	0.666
                                     \\
                                     & \textbf{GRU-D}
                                     & 0.675 &	0.686 &	0.674
                                     \\ 
                                     & \textbf{TBM}
                                     & 0.664 &	0.673 &	0.659
                                     \\ 
                                     & \textbf{BA-mean}   & 0.655  & 0.665	& 0.656   
                                     \\ \hline \hline
\end{tabular}

\caption{Precision, Recall and F-score on all datasets for all variants of FIT network (FIT, FIT-V, Multi-FIT, Multi-FIT-V), state of the art models (GRU-D, TBM), baseline model (BA-mean) and upper bound model (BA-Oracle, where applicable).}
\label{table:prf_table}
\end{table}

\subsection{PhysioNet Challenge 2012}
The PhysioNet Challenge 2012 dataset \cite{Physionet} is publicly available and contains the de-identified records of $8000$ patients in Intensive Care Units (ICU). Each record consists of roughly $48$ hours of multivariate time series data with up to $37$ features recorded at various times from the patients during their stay such as respiratory rate, glucose etc.\footnote{List of all 37 features is provided in Appendix A as Table 2.} We only used the data provided in Training Set A, which contains a subset of $4000$ patients, as this is the only set for which labels are publicly available. Researchers primarily work on predicting either in-hospital death or subsequent survival of the patient. The former being the easier task, as the data is collected when the patient is in ICU which makes it easier to predict deaths during that time period. While past work focuses on this task, we undertake the latter and the more difficult one -- post-treatment survival.
We divided the patients into two groups according to the survival index, the group which died during the scope of the study and the group which survived. This criterion divided the patients into two almost balanced groups, with $2526$ people surviving the study. The PhysioNet dataset has all the three characteristics mentioned in section~\ref{sec:introduction} which makes it a challenging dataset for the purposes of classifying surviving vs. non-surviving patients. In our experiments, we evaluated the models on 3 different splits of the data and averaged their performance over these splits. The data was divided into training, validation and testing set in a ratio of 80:10:10. For this dataset we used more training data as compared to others because of the higher difficulty of survival index prediction on this dataset.

For PhysioNet, it is not possible to train the BA-Oracle model because the values are already missing in the dataset. We obtained a lower bound of performance using BA-mean, which achieves an F-score of $0.656$. PhysioNet also has a larger feature set (37) as compared to other datasets. For this experiment we did not create another reduced dataset with fewer features. Instead, by relying on inter-feature correlations, we defined $2$ supporting features for each feature to use within FIT-V. The $2$ supporting features were chosen according to the highest absolute correlation with the relevant feature. As shown in the last line of Table~\ref{table:prf_table}, Multi-FIT-V has the best performance with an F-score of $0.681$, closely followed by FIT-V ($0.676$). GRU-D achieved an f-score of $0.674$ and FIT and Multi-FIT have f-scores of $0.666$ and $0.663$. TBM ($0.659$) and BA-mean ($0.656$) achieved the worst performance for this predictive task. The PhysioNet dataset is quite difficult to model as it has a lot of missing data (close to $84$\% for some features), making it a difficult task for all the evaluated models, which is why the baseline was difficult to improve upon even for the competitive models. The frequency of the features also vastly differs from each other. For example, heart rate is collected frequently whereas albumin concentration in blood is not. Fig. 5 in Appendix A shows the difference between three different features in terms of their data resolution. Multi-FIT-V performs better on these models since it captures the inter-correlation between different vital signals along with modeling frequent and less-frequent signals separately.

\section{Discussion and Conclusions}

Multi-resolution, irregularity and missing data pose significant problems in training classification models for multivariate time series data. By simply aligning the different signals and imputing the missing values, off-the-shelf recurrent neural networks are not able to leverage the information contained in the patterns of missing data and time intervals between data collection, and thus cannot perform well in these datasets. To address these challenges, we introduced the Flexible Irregular Time Series Network (FIT) which uses a fully connected neural network (\emph{memory cell}) to impute the missing values in datasets in the form of representations learned from missing flags, the average value of the signal and time intervals along with actual data values, in conjunction with a sequence model. We also introduced the Vertical Flexible Irregular Time Series Network (FIT-V) model, which handles missingness in a signal by also taking into account the other features correlated with it, referred to as \emph{support signals}. 

Furthermore, we introduced two other models specifically designed for multi-resolution data -- Multi-FIT and Multi-FIT-V. These models have two components, one which handles slow features and the other which handles the fast ones, the representations from both components are concatenated for label prediction. We have shown through experiments presented in Section~\ref{sec:Experiments} that, as hypothesized, FIT outperforms the baseline model BA-mean, as well as the state-of-the-art TBM on OU-ISIR and HMP along with GRU-D on PhysioNet dataset, due to its ability to learn complex representations leveraging information in the pattern of missing data and the duration between observations.

Our experiments show that FIT-V further improves predictive performance when compared to FIT, by using few support features with strong correlations. This shows that available correlations between features can be leveraged to accurately impute missing values in time series datasets, which can in turn lead to better prediction results on multivariate time series datasets. Other than correlation, domain knowledge can also help in defining the support features. We have also showed that the FIT model is quite robust to the proportion of missing data as compared to its contenders.




Additionally, we have shown through experiments on the PhysioNet dataset that Multi-FIT and Multi-FIT-V achieve improved results on datasets with multi-resolution signals. These models do not impute slow signals at each time step, leading to a sequence with fewer steps that assists the sequence model in building a meaningful representation. In PhysioNet, we also compared Multi-FIT-V's result against another state-of-the-art model Gated Recurrent Unit with decay (GRU-D) and showed that Multi-FIT-V achieves a better performance.

In conclusion, we have shown 
that Multi-FIT networks 
overcome the major problems of multi-resolution, irregularity and missing values frequently encountered in medical datasets, improving on the state-of-the-art performance for patient survival forecasting.
We also saw that FIT and FIT-V can build efficient and accurate representations for missing values on the OU-ISIR and HMP datasets, leading to considerable improvements in predictive performance.


\todo{
\begin{itemize}
    \item 
    We introduced deep belief network  that can handle irregular and multi-resolution data. We showed that our method works better than previous methods by extensive experiments on real data, some with controlled patterns of missingness
    \item Emphasis why our models are better than the previous models and why is it important to solve this problem and how it is also a problem in the other fields as well and not just medical domains.
    \item Propose the changes that we need to do in the model which this model might not have been able to capture while citing more datasets to test that hypothesis.
\end{itemize}
}

\todo{(POSTERITY) We should state some takeaway messages that go beyond our models. Basically, each hypothesis should lead to an intuition that transcends the current paper and offers something to the community. Can you guys do a first pass on this by transferring our concrete results, which I've put between parentheses () to high-level takeaways which could be useful beyond these specific models?
\begin{itemize}
\item (DBM outperforms TBM on multivariate time series which have missing values, where the pattern of missingness differs for each signal) $\Rightarrow$ high-level takeaway? \iman{do we show this? I don't think we have any proof for this, but basically if this is true, its because DBM can handle more complex patterns I think}
\item (VDBM outperforms DBM when there is feature correlation, as shown by our OU-ISIR-Corr and HMP results) $\Rightarrow$ \iman{done}\iman{This proves that correlation can be used in imputing missing data}
\item (If the feature set is large, then for VDBM one should select a couple features either on the basis of inter-corr or domain knowledge for enhanced performance.) \iman{done}\\   \iman{making model too complex basically does not work and VDBM with many features are too complex so they bring more harm than good. Maybe if there were enough data to learn this the complex model with many features would be able to assign weights such that the correlated features get more weights, however, there isn't.}\\ \iman{This is what I wrote: VDBM cannot outperform DBM as by increasing the number of parameters VDBM becomes too complex and there is not enough available data to train such a complex model. This shows that when leveraging the correlations of of features in these datasets, unless the size of dataset is really big, the features which will be used to impute each other should be selected by expert or correlation analysis.}
\item (VDBM may not outperform DBM when there are a lot of weakly correlated features as shown by our OU-ISIR vs. OU-ISIR-Corr because using a lot of weakly-correlated features may unintentionally add noise) $\Rightarrow$
\iman{done} \\
\iman{this is what I wrote: Furthermore, when there are some strongly correlated features provided for a feature, adding more features with weak correlation to it will unnecessarily increase the complexity of the model and will reduce the performance of the model.}
\item (DBM and VDBM in general need more additional changes to adequately handle data such as Physionet which have high missing data rate as well as a lot of weakly correlated features. Even TBM, as all these models are not much better than BA-Avg) \iman{done} \\
\iman{This is what I wrote: Finally, it can be seen from results achieved on PhysioNet dataset that the gap between the best model (VDBM) and the baseline (BA-Avg) is not marginal, and overall all the methods do not perform as well as other datasets here. This shows that as the dataset becomes very sparse, it is hard to select correlated features by only looking at inter-correlations, as there are basically not enough non-missing data available to infer these correlation values, in these cases, it would be better to select supporting features according to experts opinions rather than relying solely on inter-correlation values.}

\item anything else?
\end{itemize}
}


\bibliographystyle{named}

\clearpage
\section*{Appendix A}
\begin{figure}[!htbp]
\centering
  \includegraphics[width=8cm]{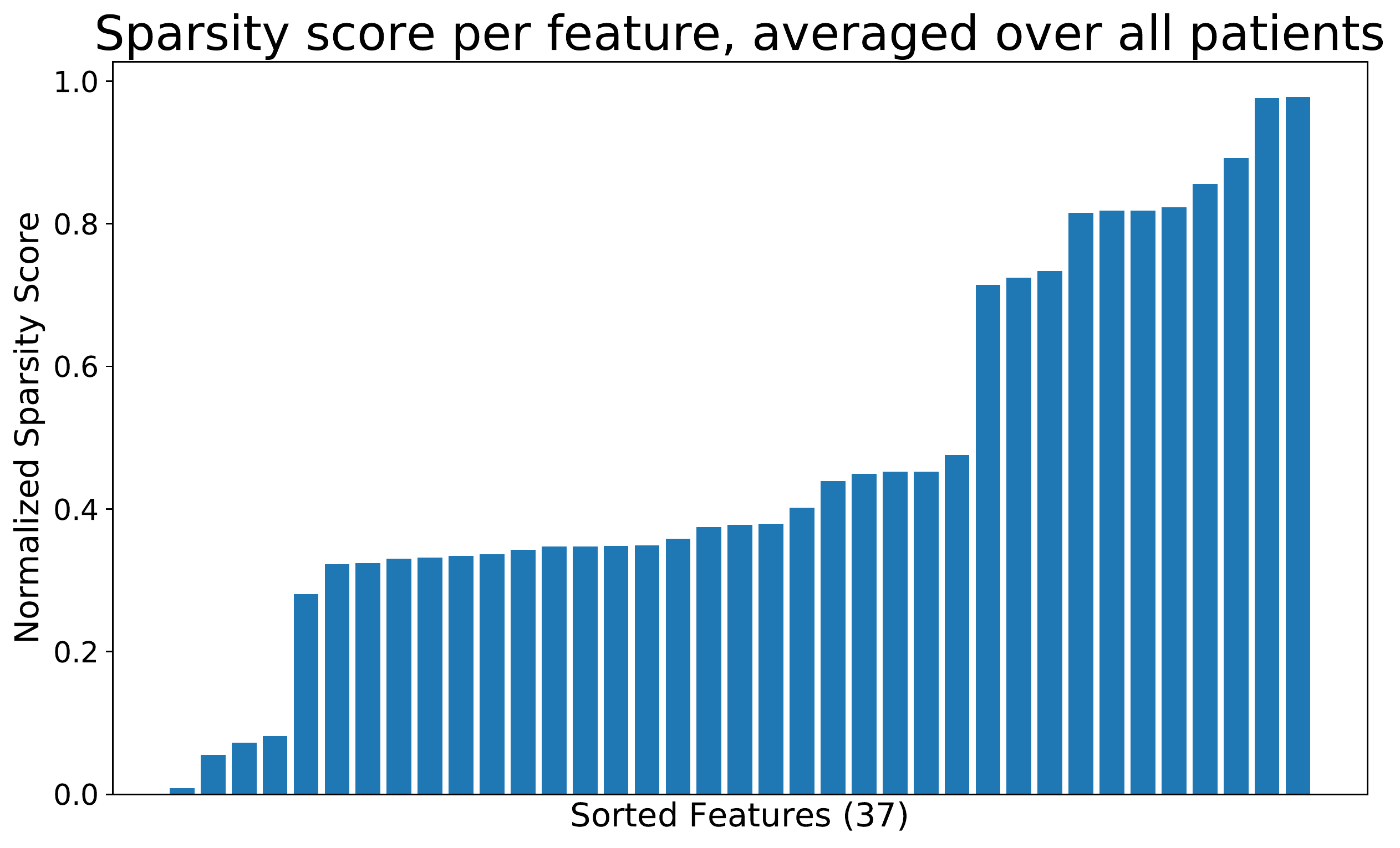}
  \caption{Sparsity score for each feature which helps in dividing each feature into three different resolution bins.}
  \label{score_fig}
\end{figure}
\begin{figure}[!htbp]
\centering
  \includegraphics[width=8cm]{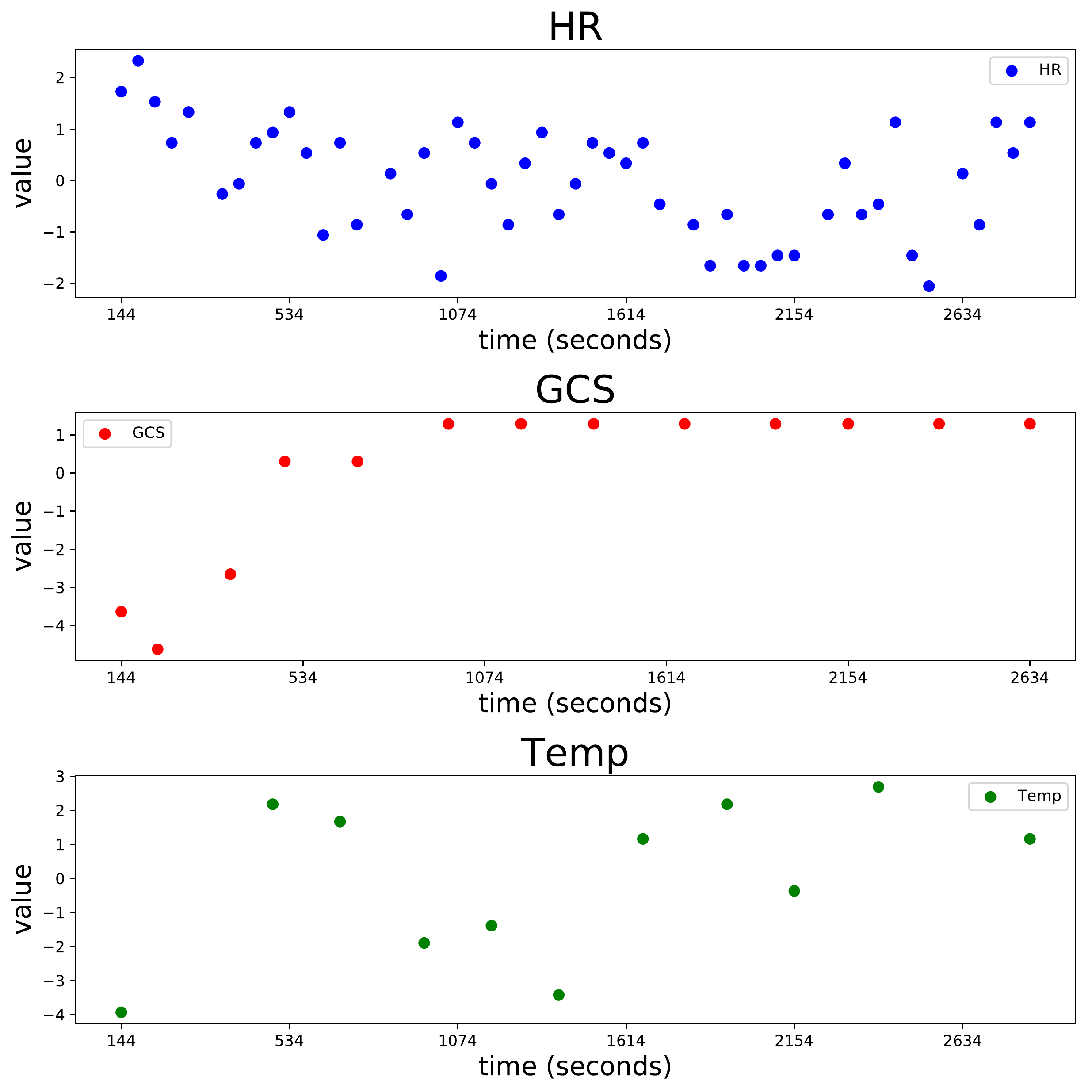}
  \caption{Three example features in Physionet dataset. HR (Heart-rate) is regular with high frequency, GCS(Glasgow Coma Score) is also regular but it has a lower frequency so HR and GCS are an example of multi-resolution and Temp (Temperature) is irregular.}
  \label{signals}
\end{figure}
\pagebreak
\begin{table}[ht!]
    \centering
\begin{tabular}{c l c} 
\toprule
 Index & Variable Name & Missing Rate \\
 \hline
1 & ALP & 0.9875\\
\hline
2 & ALT & 0.9871\\
\hline
3 & AST & 0.9871\\
\hline
4 & Albumin & 0.9903\\
\hline
5 & BUN & 0.9447\\
\hline
6 & Bilirubin & 0.9871\\
\hline
7 & Cholesterol & 0.9987\\
\hline
8 & Creatinine & 0.9445\\
\hline
9 & DiasABP & 0.5594\\
\hline
10 & FiO2 & 0.8989\\
\hline
11 & GCS & 0.7789\\
\hline
12 & Glucose & 0.9473\\
\hline
13 & HCO3 & 0.9456\\
\hline
14 & HCT & 0.9303\\
\hline
15 & HR & 0.2218\\
\hline
16 & K & 0.9417\\
\hline
17 & Lactate & 0.9752\\
\hline
18 & MAP & 0.5647\\
\hline
19 & MechVent & 0.9042\\
\hline
20 & Mg & 0.9464\\
\hline
21 & NIDiasABP & 0.6203\\
\hline
22 & NIMAP & 0.6255\\
\hline
23 & NISysABP & 0.6199\\
\hline
24 & Na & 0.9453\\
\hline
25 & PaCO2 & 0.9303\\
\hline
26 & PaO2 & 0.9303\\
\hline
27 & Platelets & 0.9459\\
\hline
28 & RespRate & 0.7682\\
\hline
29 & SaO2 & 0.977\\
\hline
30 & SysABP & 0.5592\\
\hline
31 & Temp & 0.7259\\
\hline
32 & TroponinI & 0.9984\\
\hline
33 & TroponinT & 0.9912\\
\hline
34 & Urine & 0.5355\\
\hline
35 & WBC & 0.9496\\
\hline
36 & Weight & 0.5741\\
\hline
37 & pH & 0.9272\\
\bottomrule
\end{tabular}
\caption{List of 37 features from Physionet Dataset and their corresponding missing rate}
\label{fig:my_label}
\end{table}
\end{document}